\documentclass[conference]{IEEEtran}
\usepackage{bm}
\usepackage{booktabs}
\usepackage{multirow}
\usepackage{cite}
\usepackage{amsmath,amssymb,amsfonts}
\usepackage{algorithmic}
\usepackage{graphicx}
\usepackage{textcomp}
\usepackage{xcolor}
\usepackage{comment}
\usepackage{dirtytalk}
\usepackage{hyperref}
\usepackage{booktabs}
\def\BibTeX{{\rm ``B\kern-.05em{\sc i\kern-.025em b}\kern-.08em
    T\kern-.1667em\lower.7ex\hbox{E}\kern-.125emX''}}

\usepackage{background}
\usepackage{tikz}
\usepackage{xcolor}

\backgroundsetup{
  scale=14,
  angle=45,
  opacity=0.3,
  contents={\bf\textcolor{gray}{PREPRINT}}
}
\begin{document}

\title{Multimodal Power Outage Prediction for Rapid Disaster Response and Resource Allocation\\
{\footnotesize }

}

\author{\IEEEauthorblockN{Alejandro Aparcedo\textsuperscript{1, 2, 3}, Christian Lopez \textsuperscript{1, 2, 3}, Abhinav Kotta \textsuperscript{1, 2, 3}, Mengjie Li\textsuperscript{1, 2, 3, 4}}
\\

\IEEEauthorblockA{
\textsuperscript{1} Florida Solar Energy Center (FSEC), University of Central Florida (UCF), Cocoa, FL 32922, USA\\
\textsuperscript{2} Resilient, Intelligent and Sustainable Energy Systems (RISES) Cluster, UCF, Orlando, FL 32816, USA\\
\textsuperscript{3} Department of Computer Science, University of Central Florida (UCF), Orlando, FL 32816, USA\\
\textsuperscript{4} Department of Materials Science and Engineering, UCF, Orlando, FL 32816, USA\\
}
}
\maketitle

\begin{figure*}[!t]
\centerline{\includegraphics[width=\textwidth] {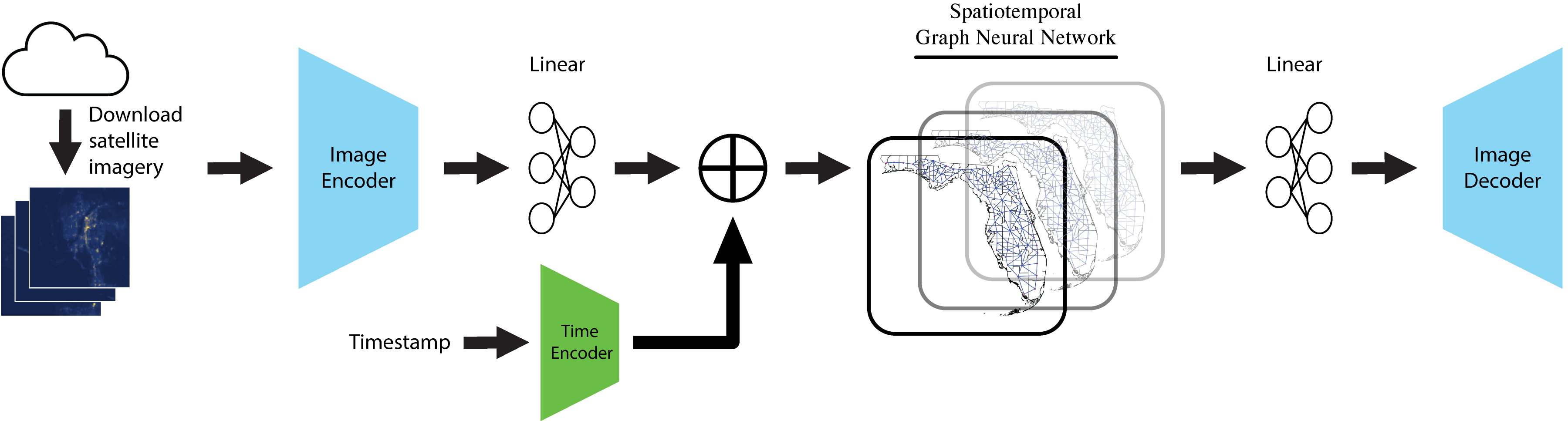}}
\caption{Workflow of Visual-Spatiotemporal Graph Neural Network for NTL and power outage prediction. First, download images from NASA Black Marble. Images are downspampled by image encoder and projected onto lower-dimensional space with a linear layer. Following, we concatenate the image embedding with a time embedding which is used as input for the st-GNN. The output of the st-GNN will be a future $T$ embedding based on $S$ past graph signals, this future embedding is projected onto higher-dimensional space and upscaled using an image decoder to obtain pixel-level predictions of NTL and power outages.}
\label{model}
\end{figure*}

\begin{abstract}
Extreme weather events are increasingly common due to climate change, posing significant risks. To mitigate further damage, a shift towards renewable energy is imperative. Unfortunately, underrepresented communities that are most affected often receive infrastructure improvements last. We propose a novel visual spatiotemporal framework for predicting nighttime lights (NTL), power outage severity and location before and after major hurricanes. Central to our solution is the Visual-Spatiotemporal Graph Neural Network (VST-GNN), to learn spatial and temporal coherence from images. Our work brings awareness to underrepresented areas in urgent need of enhanced energy infrastructure, such as future photovoltaic (PV) deployment. By identifying the severity and localization of power outages, our initiative aims to raise awareness and prompt action from policymakers and community stakeholders. Ultimately, this effort seeks to empower regions with vulnerable energy infrastructure, enhancing resilience and reliability for at-risk communities.

\end{abstract}

\hfill\allowdisplaybreaks


\section{Introduction}

The assessment of satellite imagery for large-scale weather disasters presents a critical challenge, especially in the face of increasing climate change impacts. In the USA, the impact of weather and climate disasters in 2023 was of \$92.9 billion\cite{noaa_jan23_report}. Further, the number of 1 billion dollar climate disasters that occured in the USA rose to an all-time high of 28 disasters in 2023 as reported by the National Oceanic and Atmospheric Administration \cite{2023_weather_report}. Among these disasters, hurricanes stand out as some of the deadliest, costliest, and most frequent occurrences, with their devastating effects vividly illustrated by notable examples. Consider Hurricane Michael's impact on Florida's Panhandle, rendering 1.7 million people powerless \cite{ghorbanzadeh_statistical_2020}. Hurricane Ian made landfall in southwestern Florida with costs surpassing \$112 billion in damages, making it the costliest hurricane in Florida's history \cite{h_ian_report}. Ian resulted in a total of ~3.28 million customers without power \cite{h_ian_report}! Notably, hurricane-induced flooding and wind emerge as the primary culprits behind power outages, highlighting the imperative of assessing their impact on Florida's power infrastructure.

Swift action in disaster response is essential for mitigating natural disaster impacts. Our commitment to energy equity drives efforts to provide assistance to the most vulnerable communities. These communities, characterized by social vulnerability and frequent exposure to extreme weather risks and power outages,  require enhanced adaptation and energy infrastructure improvements. Expanding photovoltaic (PV) energy production and storage capacity is a great way to improve community energy resilience, equity, and access to clean energy while still lowering costs for the consumer\cite{chan_design_2017}. 

\begin{figure}
\centerline{\includegraphics[width=0.9\columnwidth]{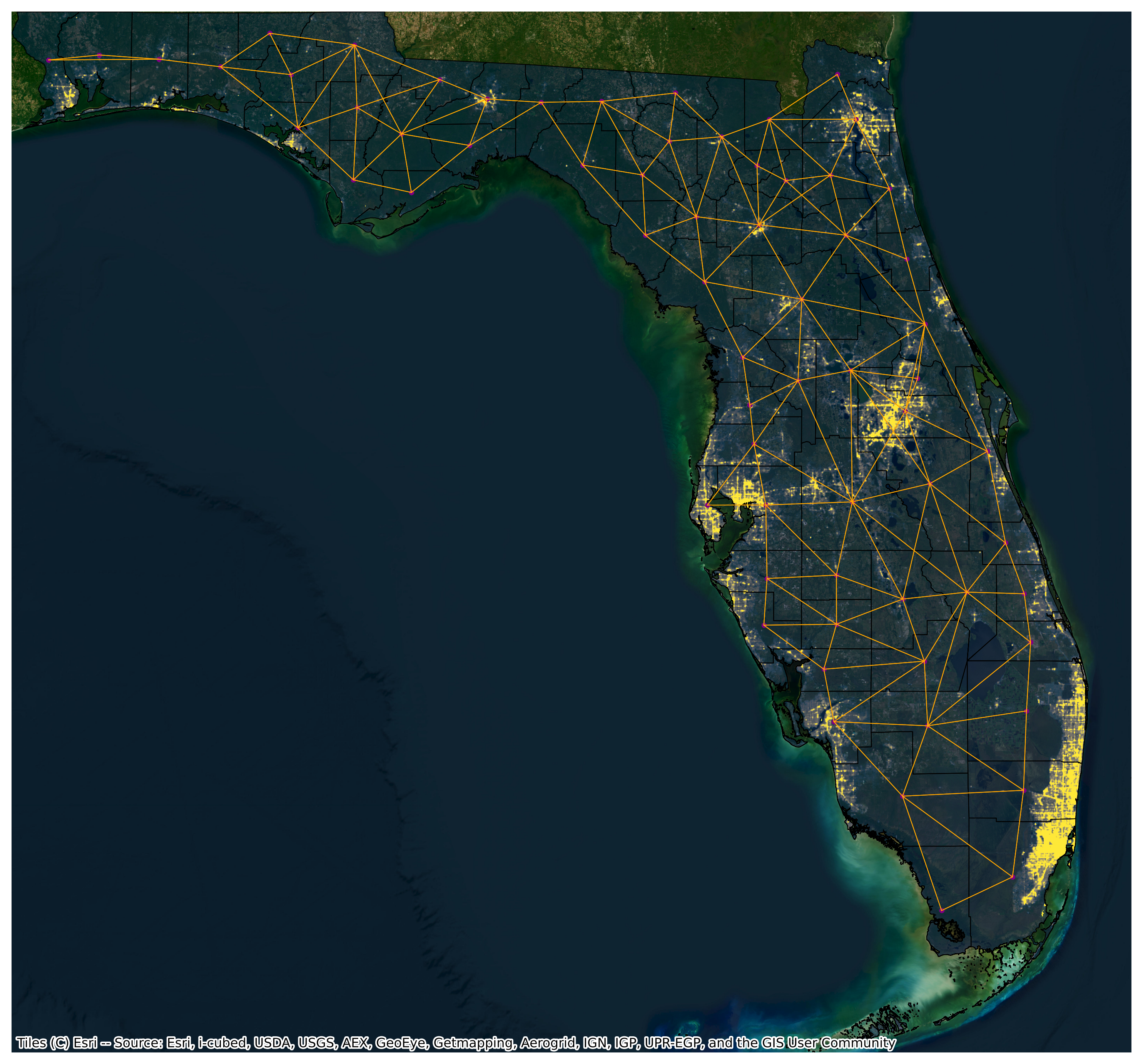}}
\caption{Visual representation of non-adaptive adjacency matrix (graph) and NTL for Florida counties. The graph is overlayed on top of a basemap provided by Esri. The NTL data is from Black Marble annual composite of 2022.}
\label{fl_graph}
\end{figure}

Nighttime satellite imagery made available by NASA's Black Marble nighttime lights (NTL) product suite provides a visual tool for identifying outages in affected communities. We propose a new architecture, a Visual-Spatiotemporal Graph Neural Network (VST-GNN) to effectively analyze night satellite imagery across diverse  geospatial regions, see Figures \ref{model} and \ref{fl_graph}. Our focus is on predicting the severity and geographic distribution of power outages, aiming to enhance disaster response and resilience.

The main contributions of our work are as follows:
\begin{itemize}
  \item Novel framework to predict nighttime lights and power outage severity and location.
  \item Empirical evaluation on the effectiveness of the proposed framework.
  \item Verify the robustness by evaluating on out-of-distribution data from other hurricanes.
\end{itemize}

\section{Related Work}

\subsection{Deep Learning in Photovoltaics}

One field that demonstrates the need for deep learning is electroluminescence (EL) image inspection. Fioresi et al. shows deep learning models can be used to identify and segment defects in PV cells from EL images \cite{fioresi_automated_2022}. Further, prediction of PV power output is a crucial task for PV module maintenance and monitoring. Multiple works have used deep learning models such as Long-Short Term Memory and Artificial Neural Networks to forecast PV power output \cite{luo_deep_2021, scott_machine_2023}. Spatiotemporal Graph Neural Networks (st-GNN) have also been used to predict the power output of PV modules \cite{karimi_spatiotemporal_2021} as well as estimating fleet-wide performance degradation \cite{fan2024_stgnn_degradation}. 
\label{deep_learning_pv}

\subsection{Nighttime Imagery}

Cao et al. proposed using S-NPP/VIIRS Day/Night Band radiance to study the effect of Hurricane Sandy and the 2012 North American derecho on the power grid of Midwest/East United States \cite{cao_detecting_2013}. Wang et al. demonstrated that the Black Marble product could be effectively employed to quantify power outages and identify regions in need of disaster relief during Hurricane Sandy and Hurricane Maria \cite{wang_monitoring_2018}. Cole et al. took a synergistic approach by combining NTL, LandScan, and census data, and training an artificial neural network to detect power outages across the continental United States using Hurricane Sandy as the case study \cite{cole_synergistic_2017}. Montoya-Rincon et al. utilize different types of data such as NTL, weather, geographical, and census data to study power outages caused by Hurricane Irma and Maria \cite{montoya-rincon_use_2022}. They test various machine learning models such as Bayesian Additive Regression Trees, Random Forest, and Extreme Gradient Boosting \cite{montoya-rincon_use_2022}. Cui et al. propose a power outage detection model that calculates the percentage of outage by calculating the average days before winter storm Uri and subtracting the radiance the days after the storm \cite{cui_disaster-caused_2023}.

\subsection{Spatial Temporal Graph Neural Networks}

Graph Neural Networks (GNNs) have demonstrated their effectiveness in learning from interdependent data, with successful applications in fields such as weather forecasting \cite{graphcast} and traffic system prediction \cite{li2023graph}. However, many tasks, like weather forecasting, necessitate the integration of both temporal data from previous time steps and spatial data to accurately capture the features at each time step. Graph WaveNet utilizes an adaptable adjacency matrix and a stacked dilated 1D convolution component to learn a function that is able to forecast future graph signals across spatial and temporal dimensions \cite{wu_graph_2019}. In addition, Graph WaveNet performs forecasting for the desired number of future time steps non-recursively, allowing it to outperform other current models during inference \cite{wu_graph_2019}. 

\subsection{U-Net / Visual node-level features}

U-Net, introduced by Ronneberger et al., is able to perform precise localization on an image by utilizing contracting and expanding components. The contracting path down-samples the input image while simultaneously extracting and storing significant features. The expansive path subsequently up-samples the input using convolutions that incorporate the previously stored features. The result is a high-resolution image generated pixel-by-pixel, where each pixel's value is determined based on its surrounding context, making them particularly suitable for pixel-level segmentation in medical images \cite{ronneberger_u-net_2015}. U-Nets are based on Convolutional Neural Networks (CNNs) \cite{lecun1998} and skip connections which aid in learning of image features. These connections concatenate the feature maps produced before certain convolutional layers with those in the expansive portion of the network. This preserves essential visual features and contributes to creating higher-quality segmented images \cite{wilm2024rethinking}.

As GNNs have an extensive range of applications, researchers have focused on combining U-Net with GNNs to enhance the handling of graph-structured data. Hermes et al. replaced the convolutions in the original U-Net architecture with GNN operations, showing that the graph pooling operations can be advantageous when predicting traffic in unseen cities \cite{hermes_graph-based_2022}. This modification has proven to be helpful for traffic forecasting \cite{gao_graph_2022}. These works have focused on contracting and expanding the graph itself by using graph-specific convolutions rather than utilizing existing convolutional operations to reduce the dimensionality of visual input features.

\section{Methodology}
\subsection{Problem Definition}
In this section, we go into detail of our proposed VST-GNN. While deep learning and the st-GNN architecture has been used in photovoltaics before in various areas, see \ref{deep_learning_pv}, our work is mainly motivated by \cite{karimi_spatiotemporal_2021}. Focusing on predicting the severity of power outages in the state of Florida, we consider each county as a node in a graph, $G = (V, E)$, where $V$ is the set of nodes and $E$ is the set of edges. Let $|V|$ denote the total number of nodes (counties). Given a graph $G$, historical S step graph signals, and an input image sequence $X \in \mathbb{R}^{C \times H \times W}$, where $C$ is the number of channels, $H$ is the height and $W$ is the width of each image, our problem is to learn a function $f$ which is able to forecast its next $T$ step graph signals. We can formulate our problem as:
\begin{equation}
X^{(t-S):t} \xrightarrow{f(\cdot)} X^{(t+1):(t+T)}
\end{equation}
where $X^{(t-S):t} \in \mathbb{R}^{|V| \times S \times C \times H \times W}$ and $X^{(t+1):(t+T)} \in \mathbb{R}^{|V| \times T \times C \times H \times W }$.
\subsection{Model Architecture}
We propose deriving graph signals from Black Marble satellite imagery, with historical $S$ step graph signals and nodes $V$. The image will be downsampled by the U-Net image encoder, flattened, and projected onto a lower-dimensional space using a linear layer to produce our image embedding $\textbf{v}$, denoted by $g(\cdot)$.
\begin{equation}
    X^{(t-S):t} \xrightarrow{g(\cdot)}  {\textbf{v}}^{(t-S):t}
\end{equation}
where ${\textbf{v}}^{(t-S):t} \in \mathbb{R}^{|V| \times S \times P}$, $P$ being the size of the image embedding. We concatenate our image embedding with a time embedding from \cite{kazemi_time2vec_2019}, which we will denote as $\bm{\tau}$.
\begin{equation}
{\textbf{z}}^{(t-S):t} \leftarrow concat({\textbf{v}}^{(t-S):t}, {\bm{\tau}}^{(t-S):t})
\end{equation}
${\textbf{z}}^{(t-S):t} \in \mathbb{R}^{|V| \times S \times Z}$, where $Z$ is the dimension of the combined image and temporal embedding. ${\textbf{z}}^{(t-S):t}$ is processed with a st-GNN $h(\cdot)$ to learn spatial and temporal coherence between graph signals, producing an image embedding for a future graph signal $T$.
\begin{equation}
{\textbf{z}}^{(t-S):t} \xrightarrow{h(\cdot)} {\textbf{v}}^{(t+1):(t+T)}
\end{equation}
where ${\textbf{v}}^{(t+1):(t+T)} \in \mathbb{R}^{|V| \times T \times P}$. Finally, the future embedding is projected onto higher-dimensional space using a linear layer, reshape, and upscaled using an image decoder, denoted as $u(\cdot)$.
\begin{equation}
{\textbf{v}}^{(t+1):(t+T)} \xrightarrow{u(\cdot)} X^{(t+1):(t+T)}
\end{equation}
An adaptive adjacency matrix, representing the set of edges $E$, is learned as a parameter during st-GNN training through backpropagation \cite{wu_graph_2019}. This allows the model to automatically discover and optimize the graph structure. A visual representation of the initial non-adaptive adjacency matrix can be seen in Figure \ref{fl_graph}.

\subsection{Datasets}

\textbf{NASA’s Black Marble.} The Black Marble dataset contains data from the Visible Infrared Imaging Radiometer Suite (VIIRS) Day/Night Band (DNB) onboard the Suomi National Polar-orbiting Platform \cite{roman_nasas_2018}. The dataset is available in raster format with a spatial resolution of 15 arc seconds or approximately 500 meters and a time resolution of daily (VNP46A2), monthly (VNP46A3), or annual (VNP46A4). The data is measured in nanowatts per square centimeter per steradian $nW/cm^{2}sr$. We download images 30 days before and after each major weather event using the BlackMarblePy package\cite{blackmarblepy}. Each county is represented by a bounding box around the county region of interest. We do not remove any low quality or gap filled observation during the download. Pixel values representing the ocean are set to a fill value, these pixels are set to zero when the data is loaded before training. 

To quantify the severity of a power outage we follow \cite{cole_synergistic_2017, wang_monitoring_2018, cui_disaster-caused_2023}, 

\begin{equation}
    Percent_{Normal} = 100 \times \frac{NTL}{NTL_{Normal}}
\end{equation}

where $NTL$ represents the radiance of any given day and $NTL_{Normal}$ represents the average of the last three available monthly composites (i.e., Black Marble product VNP46A3) for that day.  

\section{Experiments}

For our experiments we focus on three major hurricanes that have occurred in Florida in recent years. We study Hurricane Michael (H-Michael), Ian (H-Ian), and Idalia (H-Idalia) and their effect on the power energy grid infrastructure of Florida. 

\subsection{Case Studies}

\begin{figure*}[!t]
\centerline{\includegraphics[width=\textwidth] {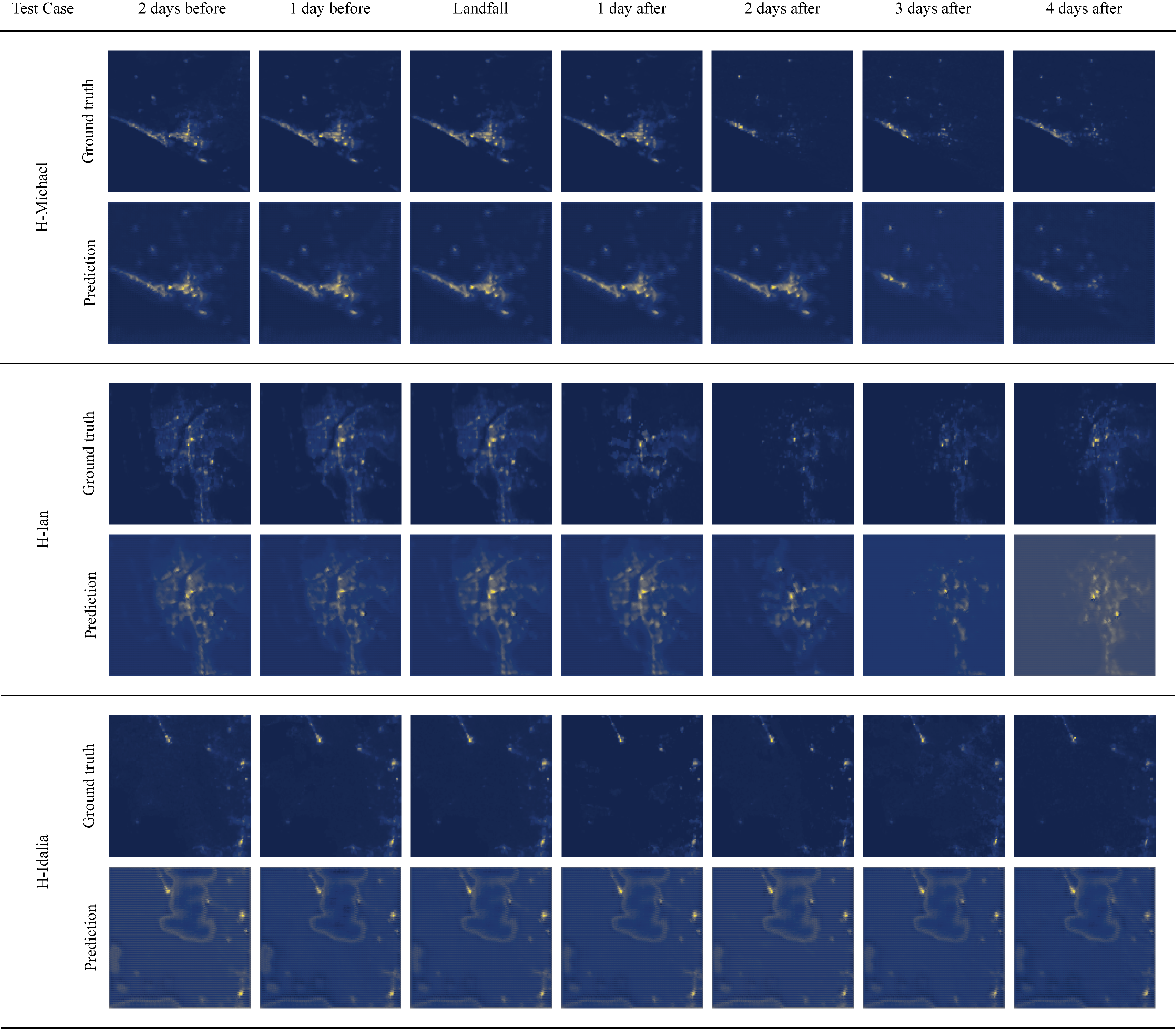}}
\caption{Results from evaluating on a single hurricane at a time. Testing on H-Michael (Bay County) and H-Ian (Lee County) demonstrates power outage prediction effectiveness once the power outage has happened (after landfall) and accurately predicts areas where power outages occur. Subsequently, following the hurricane, the model is able to predict the areas whose power was recovered first. The last two rows shows a failure case, H-Idalia (Levy County), where the model correctly identifies the areas with most light but also generates incorrect nightlight patterns. Note, each test is out of distribution as the model was only trained on data from the other two hurricanes.}
\label{nightlights}
\end{figure*}

\begin{figure*}[!t]
\centerline{\includegraphics[width=\textwidth] {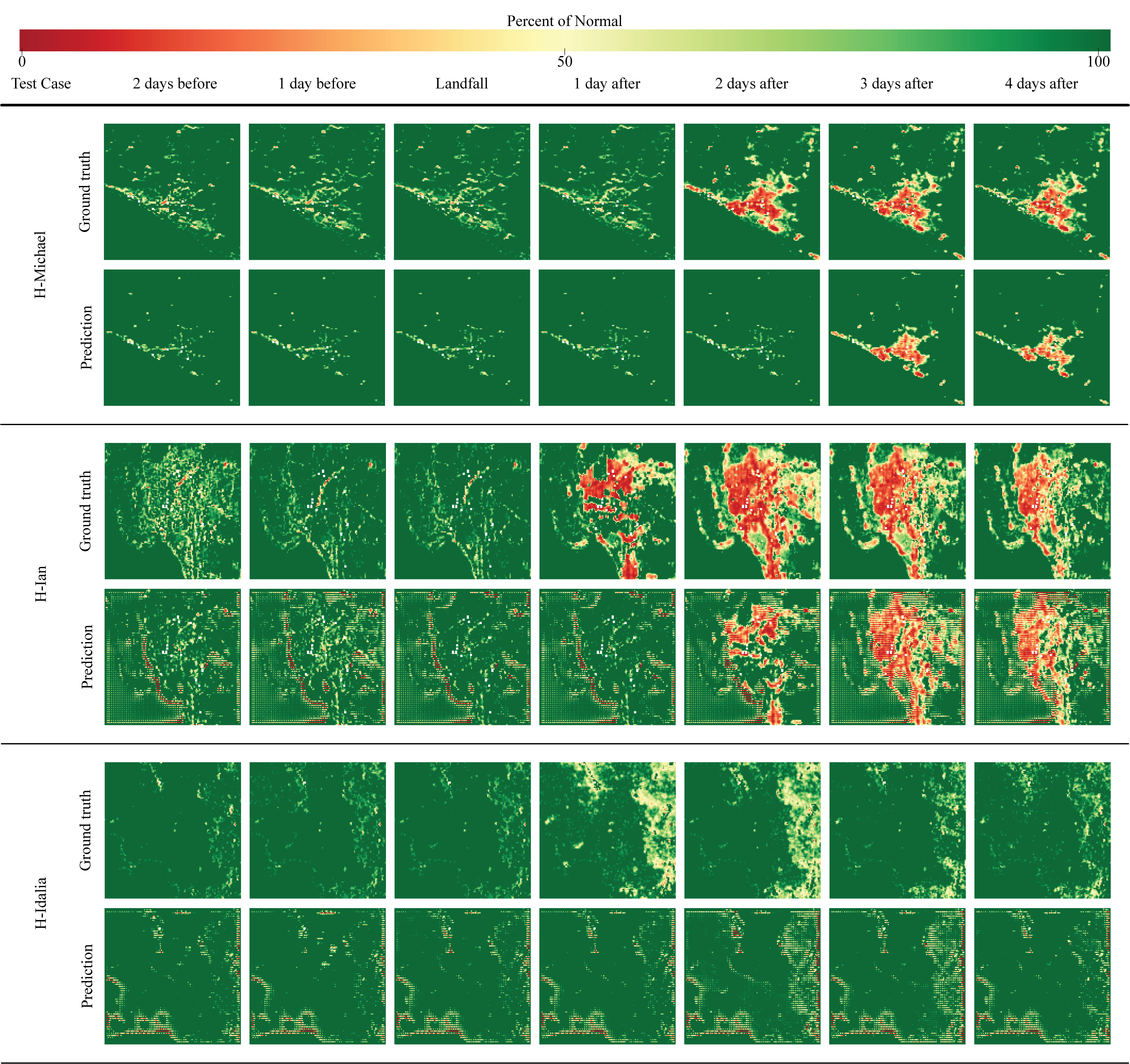}}

\caption{Results from our case studies, evaluating on a single hurricane at a time. Each image from Figure \ref{nightlights} was processed to produce a Percent of Normal map. Consistent with Figure \ref{nightlights}, we see for H-Michael (Bay County) the predictions are sufficient to quantify the severity and localization of power outages. For H-Ian (Lee County), severity and recovery predictions are accurate but we see higher levels of noise in the output. The bottom two rows show our failure case, H-Idalia (Levy County) exhibiting similar levels of noise to the H-Ian case. The color represents the severity of the outage, ranging from 0\% (red, severe outage) to 100\% normal (green, no outage).}
\label{risk_maps}
\end{figure*}

In this section, we present our results around three major case studies: H-Michael, H-Ian, H-Idalia. Results are presented as follows:
\begin{itemize}
    \item Case Michael: train/validation on H-Ian and H-Idalia; test on H-Michael.
    \item Case Ian: train/validate on H-Michael and H-Idalia; test on H-Ian.
    \item Case Idalia: train/validate on H-Michael and H-Ian; test on H-Idalia.
\end{itemize}

Quantitative results are shown in Table \ref{tab:results}. Qualitative results are illustrated in Figure \ref{nightlights} and Figure \ref{risk_maps}. For each case, we evaluate using out-of-distribution data: two hurricanes' data are split into training and validation sets, while the third hurricane's data is used for testing. Our results in Figure \ref{nightlights}  show that our VST-GNN model is able to accurately predict NTL patterns before and after the hurricane. The predictions after the hurricane landfall seem to be one timestep behind, intuitively, this could be because there is no data indicating the potential of a hurricane occurring so it has no way of knowing exactly when it will occur. Nonetheless, once the power loss occurs, the model is able to predict the severity and location of outages for the following days, see \ref{nightlights} and \ref{risk_maps}. Further, our quantitative results \ref{tab:results} show the robustness of our model when tested on different hurricanes. 

\begin{table}[h!]
    \centering
    \begin{tabular}{c c c c}
        \hline
        Case & RMSE & MAE & MAPE \\ \hline
        Michael & 0.43 & 0.20 & 146.94\% \\ 
        Ian & 0.40 & 0.18 & 151.53\% \\ 
        Idalia & 0.40 & 0.18 & 153.45\% \\ \hline
    \end{tabular}
    \vspace{0.2cm} 
    \caption{Test results on each case study. Each case is trained and validated on data from the other two hurricanes.}
    \label{tab:results}
\end{table}

\subsection{Setup}

Our VST-GNN model is trained end-to-end with a batch size of 16 and a random seed of 42. The raster data is resized to size 128x128 on model ingestion. We use the ADAM optimizer \cite{kingma_adam_2015} with a learning rate of 0.001, cosine annealing learning rate schedule \cite{loshchilov_sgdr_2017_cosine_annealing}, and train with mean squared error loss. To map spatiotemporal dependencies we use off-the-shelf Graph WaveNet \cite{wu_graph_2019} with a horizon of 1 timestep. Size of training, validation, and testing sets is 83, 35, and 53, respectively. Our model is trained and evaluated using PyTorch Lighting. Each raster image is compressed by the image encoder and projected to create an image embedding of size 256. The image embedding is concatenated with the time embedding from \cite{kazemi_time2vec_2019} of size 64 to create feature vector size of 320. We report mean absolute error (MAE), root mean squared error (RMSE), and mean absolute percentage error (MAPE). All experiments were done on a single NVIDIA H100 with 80 GB of memory.


\section{Discussion}
In this work, we combine visual feature extractor and st-GNN models based on the motivation that adjacent counties will be similarly affected by a major hurricane and that this effect will be visible from Earth orbiting satellites. This combined approach has proven effective in predicting power outages caused by hurricanes in Florida, as evidenced by previous work and our own results. Figures \ref{nightlights} and \ref{risk_maps} show the effectiveness of our proposed framework, despite some level of error, in identifying the severity and localization of power outages in 2/3 of our test cases. Further, Table \ref{tab:results} shows that our VST-GNN framework is robust when evaluated with data from various hurricanes.

Based on our results, we conjecture our methodology can be generalized to predict NTL and power outages from other weather-related events, thus demonstrating its broader applicability and potential. Future work could enhance this strategy by incorporating diverse data types, such as weather, demographic, and topographical information, to further boost model performance.

\section{Conclusion}
In this work, we propose a novel deep learning framework to predict the severity and localization of NTL and power outages before and after major weather events. We evaluate our method on three separate major hurricane events in the state of Florida. Our results indicate that the proposed VST-GNN model is effective and robust in identifying the severity and localization of NTL and power outages after major hurricane events in Florida. We hope our work brings awareness to underrepresented communities that consistently experience power outages caused by major weather events. With the goal of reaching 100\% renewable energy in mind, we propose underrepresented areas may be assisted by improvements in infrastructure such as PV system deployment along with energy storage to reduce cost and improve energy grid resilience.

\section*{Acknowledgment}
The authors would like to acknowledge Dr. Kristopher Davis for thoughtful discussion and Elizabeth Trader for initial data collection. 
The work is supported by DOE funded project CARES under DE-EE0010418. 
Support for DOI 10.13139/ORNLNCCS/1975203 dataset is provided by the U.S. Department of Energy, project EAGLE-I under Contract DE-AC05-00OR22725. Project EAGLE-I used resources of the Oak Ridge Leadership Computing Facility at Oak Ridge National Laboratory, which is supported by the Office of Science of the U.S. Department of Energy under Contract No. DE-AC05-00OR22725

\nocite{*} 
\bibliographystyle{ieeetr}
\bibliography{references}
\end{document}